\pdfoutput=1
\documentclass[11pt]{article}

\usepackage[final]{acl}
\usepackage{graphicx}
\usepackage{times}
\usepackage{latexsym}
\usepackage{enumitem}
\usepackage[T1]{fontenc}
\usepackage{makecell}

\usepackage[utf8]{inputenc}

\usepackage{microtype}

\usepackage{inconsolata}

\title{Cheap Ways of Extracting Clinical Markers from Texts}

\author{Anastasia Sandu \\ \texttt{\fontsize{11}{11.5}\selectfont anastasiasandu777@gmail.com} \And 
        Teodor Mihailescu \\ \texttt{\fontsize{11}{11.5}\selectfont teomihailescu@yahoo.com} \And 
        Sergiu Nisioi \\ \texttt{ \fontsize{11}{11.5}\selectfont sergiu.nisioi@unibuc.ro} \AND
        \\
        Human Language Technologies Research Center \\
        Faculty of Mathematics and Computer Science \\
        University of Bucharest }

\begin{document}
\maketitle
\begin{abstract}
%Paragraph 1: This document is a supplement to the general instructions for *ACL authors. It contains instructions for using the \LaTeX{} style files for ACL conferences. The document itself conforms to its own specifications, and is therefore an example of what your manuscript should look like. These instructions should be used both for papers submitted for review and for final versions of accepted papers.

This paper describes the work of the UniBuc Archaeology team for CLPsych's 2024 Shared Task, which involved finding evidence within the text supporting the assigned suicide risk level. Two types of evidence were required: highlights (extracting relevant spans within the text) and summaries (aggregating evidence into a synthesis). Our work focuses on evaluating Large Language Models (LLM) as opposed to an alternative method that is much more memory and resource efficient. The first approach employs a good old-fashioned machine learning (GOML) pipeline consisting of a tf-idf vectorizer with a logistic regression classifier, whose representative features are used to extract relevant highlights. 
The second, more resource intensive, uses an LLM for generating the summaries and is guided by chain-of-thought to provide sequences of text indicating clinical markers.
%\red{\textbf{Content Warning:} to illustrate examples from the dataset, language is used which may be triggering.}
\end{abstract}

\section{Introduction}
%Paragraph 1: Motivation. At a high level, what is the problem area you are working in and why is it important? It is important to set the larger context here. Why is the problem of interest and importance to the larger community? 
Suicidal-themed messages on social media platforms can represent an indicator of suffering and mental health issues.
According to \citet{harmer2022suicidal}, 6\% of individuals aged 18-25 responded affirmatively to the survey questions on suicide ideation. 
% memon2018 e dubios din pct de vedere tehnic
%Furthermore, \cite{memon2018role}, in the last 10 years, 24\% of adolescents are connected to the Internet, implying on a global scale exploration of various platforms and diverse posts covering topics ranging from news and scientific topics to instances of bullying and suicide-related content. %This matter is of importance since it provides a means of developing models capable of screening posts containing influential content. %In this context, the objective is to prevent such content from impacting an individual's cognitive processes.
%Paragraph 2: What is the specific problem considered in this paper? This paragraph narrows down the topic area of the paper. In the first paragraph you have established general context and importance. 
% cred ca scopul nu e sa blocam pe saracii oameni care sufera, ci sa identificam early ca sufera si sa venim cu metode intr-ajutor; altfel din cauza stigmei, oamenii nu prea recunosc motivele suferintei iar daca-i blocam nu mai putem sti de ei
Interdisciplinary work on psychology and computational linguistics \cite{zirikly-etal-2019-clpsych,uban2022explainability} uses statistical models to identify various risks based on the content of social media posts or based on multi-modal characteristics such as time of post, user gender and class \cite{yang2022characteristics}.
% gandurile suicidare apar si fara postari pe social media, nu cred ca influenteaza f mult
%If measures to filter or block posts with dark content are restricted, they do not propagate, implicitly sparing the individual from emotional distress, preventing feelings of compassion, sadness, or existential questioning\cite{costanza2019meaning}.
Gaining awareness of the risk of suicide is essential, as it allows state organizations to offer support to those in need, and consequently, preventive measures can be taken, potentially saving the lives of those contemplating suicide.
% tre sa fim atenti sa nu creem si noi stigma spunand ca posteaza chestii macabre
%and those influenced by these macabre posts.
Therefore, it may be beneficial from multiple perspectives to develop methods through which the presence of suicidal thoughts can be determined on the basis of text posts on social networks. However, as \citet{rezapour2023contextual} suggests, relying solely on algorithmic methods can introduce biases, risks, and, ultimately, case-by-case analyses must be carried out by experts.

%This article \cite{harmer2020suicidal} illustrates the volatility of the human mind in response to such types of posts and how complex an individual can appear on the surface. However, it contradicts this notion by revealing statistics that indicate 6\% of individuals aged 18-25 responded affirmatively to the survey question, "At any time in the past 12 months, did you seriously think about trying to kill yourself?".
%Paragraph 3: "In this paper, we show that ...". This is the key paragraph in the intro - you summarize, in one paragraph, what are the main contributions of your paper given the context you have established in paragraphs 1 and 2. What is the general approach taken? Why are the specific results significant? This paragraph must be really really good. If you can't "sell" your work at a high level in a paragraph in the intro, then you are in trouble. As a reader or reviewer, this is the paragraph that I always look for, and read very carefully. 
\begin{figure*}[htb]
    \centering
    \includegraphics[width=0.8\linewidth]{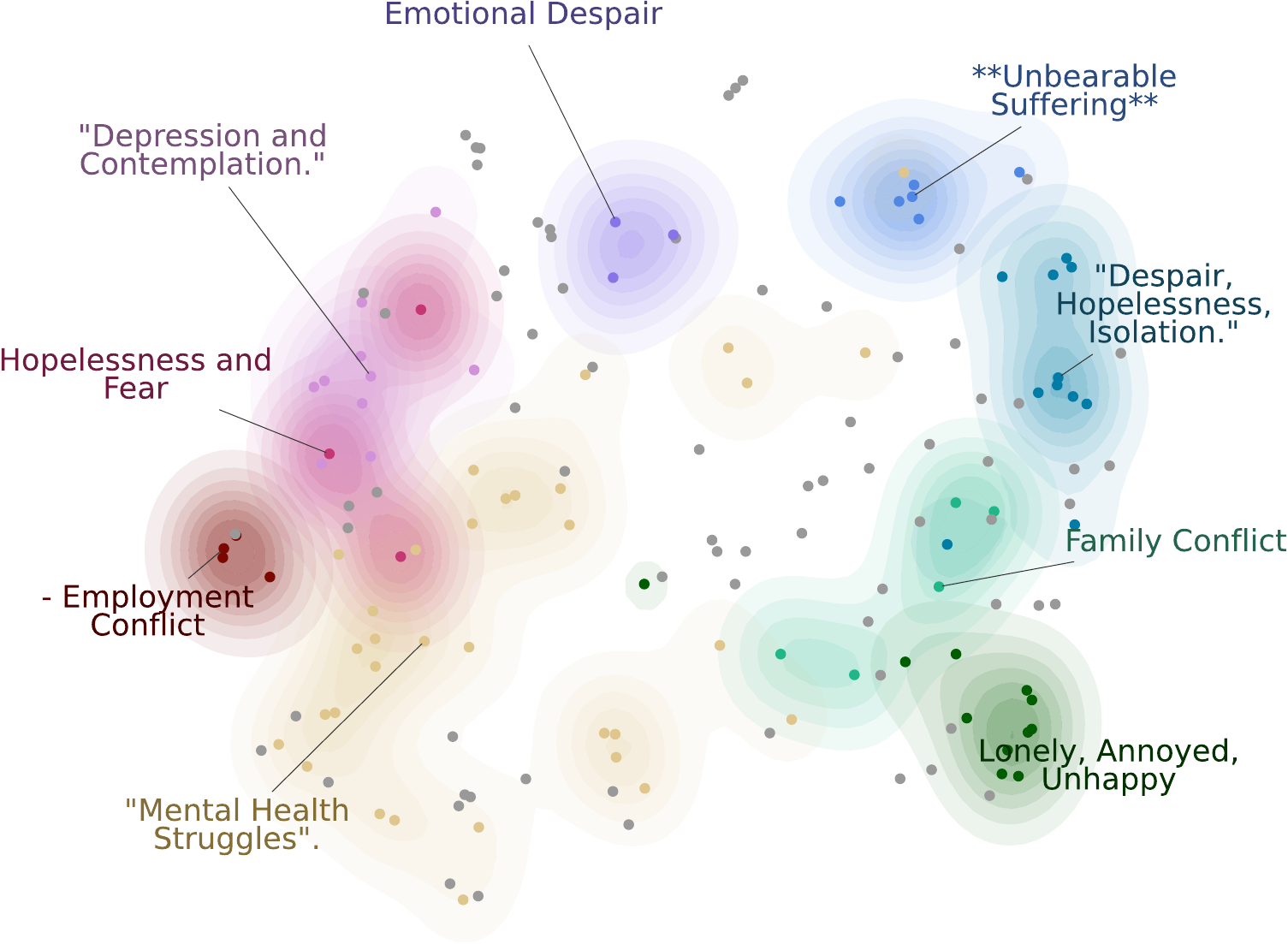}
    \caption{Major topics extracted from expert data labeled with openhermes-2.5-mistral-7b-q4\_k\_m.}
    \label{fig:topic}
\end{figure*}
% nu cred ca demonstram ceva nou, doar confirmam o gramada de studii
In this paper, as part of the shared task of the 2024 Workshop on Computational Linguistics and Clinical Psychology \cite{chim2024overview}, we address the identification of suicidal evidence in users' posts on Reddit by extracting phrases, expressions, key-words, and various types of summaries that can explain such labels. The shared task has been framed from the perspective of large language models (LLMs) with a suggestive title in this sense: "Utilising LLMs for finding supporting evidence about an individual’s suicide risk level". Although LLMs are the current standard in natural language processing \cite{mccoy2023much,fedorenko}, deploying such models at scale can be prohibitively expensive, while the pre-trainig can often be resource- and data-intensive, making such models available only for well-resourced languages and large research laboratories.

We address this task from the perspective of finding solutions for fast inference, and propose two variants: 1) to create a straightforward and \textit{cheap} (as in time-efficient) pipeline for training and identifying suicidal evidence and 2) to use prompting with quantized LLMs \cite{dettmers2023qlora} executed locally on CPU. The former is based on traditional machine learning classification techniques consisting of a tf-idf vectorizer over word ngrams paired with a feature importance selection process from a linear logistic regression classifier.

%In light of the change of AI paradigm from machine learning to deep learning, similar to the shift from symbolic logic to pattern recognition 
%In the spirit of \citet{haugeland1989artificial} who coined the term good old-fashioned artificial intelligence (GOFAI) to refer to the symbolic paradigm that has been replaced by machine-learning paradigm, we name the traditional machine learning approach as good old-fashioned machine learning.
Our results in the shared task show that a machine learning pipeline can achieve competitive evaluation scores (top 3 recall) by leveraging the risk assement annotations from the provided dataset \cite{shing-etal-2018-expert, zirikly-etal-2019-clpsych}. However, our best-performing model is a combination of LLMs used to generate good-quality summarizations and machine learning to detect highlights.

%by using the importance of the features to detect the highlights and extractive summarization to generate additional evidence. 

%Paragraph 4: At a high level what are the differences in what you are doing, and what others have done? Keep this at a high level, you can refer to a future section where specific details and differences will be given. But it is important for the reader to know at a high level, what is new about this work compared to other work in the area.

%-GML_TYPE_Q6_K - "type-0" 6-bit quantization. Super-blocks with 16 blocks, each block having 16 weights. Scales are quantized with 8 bits. This ends up using 6.5625 bpw
%LLAMA_FTYPE_MOSTLY_Q4_K_M - uses GGML_TYPE_Q6_K for half of the attention.wv and feed_forward.w2 tensors, else GGML_TYPE_Q4_K

%\begin{itemize}
%    \item the suicide risk is relatively straight forward to observe and there are little subtelties
%    \item what else?
%    \item topics
%    \item speciffic words of each class (using shifterator)
%    \item perplexity of model on texts
%\end{itemize}

\section{Data Analysis}

The annotated data provided for the shared task participants is identical to the previous edition CLPsych 2019 Shared Task: Predicting the Degree of Suicide Risk in Reddit Posts \cite{shing-etal-2018-expert,zirikly-etal-2019-clpsych} and here we include a brief summary of its subdivisions: Task A: users on r/SuicideWatch Reddit annotated based on their risk level across multiple posts using crowd-sourced annotations. Expert: user posts annotated by experts of different specialties. Tasks B and C of annotations that we did not use in this work.

All data annotations contain suicide risk categories \cite{corbitt2016college} marked with letters signifying different degrees: (a) no risk, (b) low, (c) moderate, and (d) severe risk. %Experts have a higher agreement between the annotators (0.8) than between crowd-sourced annotations (0.5), but the former do not include samples with category (a) no risk. 
Expert data is of higher quality, it consists of 332 posts, the majority (49\%) are labeled medium risk, followed by 28\% high risk and 23\% low risk. The 2024 Shared Task \cite{chim2024overview} evaluation data (not released to participants) contains additional annotations of suicide risk evidence (highlights and summaries) for 125 users of the expert subset. Our work only uses Task A and the expert subsets.

\subsection{Topic Modelling} 
%can be an effective tool 
To have a first glance over the expert-annotated data, we use the BERTopic library \cite{grootendorst2022bertopic} and embed the documents with \url{BAAI/bge-small-en} a pre-trained English model \cite{bge_embedding} which has the advantage of being relatively small and achieving good performance on the MTEB benchmark \cite{muennighoff2022mteb}. 
All document embeddings are projected into a bi-dimensional plane using a 5-neighbour UMAP \cite{mcinnes2018umap-software} configured to optimize the cosine similarity. The representations are clustered using HDBSCAN \cite{McInnes2017} with a minimum cluster size of four.
In a typical BERTopic pipeline, the topics are extracted using cTF-IDF and further fine-tuned using a representation model from openhermes-2.5-mistral-7b-q4\_k\_m\footnote{\url{https://huggingface.co/TheBloke/OpenHermes-2.5-Mistral-7B-GGUF}}. The representation model is prompted with the following statement: \textit{I have a topic that contains the following documents: [DOCUMENTS]. The topic is described by the following keywords: '[KEYWORDS]'. As an expert psychologist and therapist, provide a brief 5 word phrase to summarize the reason:}. %and the entire pipeline runs in approximately 10 minutes. 

\autoref{fig:topic} shows a result of this process with documents grouped by topic. Several key phrases are extracted using LLM prompts.
%Multiple runs tend to generate different keyword results depending on random initialization. Here, we present the result with a high diversity of topic summaries. 
Upon close inspection, the main topics in the dataset revolve around feelings of \textit{despair}, \textit{hopelessness}, socioeconomic hardships, and family conflicts. Our brief analyses indicate that the texts contain strong signals for suicide and that very few subtleties can be observed in the assessment of risk degrees.

%\section{Approaches}
%
\section{Good Old-fashioned Machine Learning (GOML)}
\label{sec:goml}
The first approach, which also obtained the highest recall amongst submissions, is based on the following steps.

\hfill \break
\textbf{1. Begin with Task A} crowd-annotated data and map the labels to binary, i.e., assigning the label 'a' to the value -1, and the labels 'b', 'c', and 'd' to the value +1.
We fit a scikit learn logistic regression classifier on tf-idf features \cite{scikit-learn}. Tokenization is done using a regular expression of the form \verb|r'\b[^\d\W]+\b'| and we employ a range of n-grams between 2 and 4 words. %Additional parameters include stripping accents, no limit on maximum number of features, and smoothing the IDF weights. 
We cross-validate several models on different subsamples of risk annotations labeled as follows: \textbf{1.1 Test} - a model trained solely on Task A test set (186 posts), \textbf{1.2 TaskA} a model trained on the entire Task A, and \textbf{1.3 A+E} a model trained on both expert and TaskA data. \autoref{tab:classification} in the appendix contains the 5-fold cross-validation results that show relatively poor classifier performance.

%Logistic Regression is a machine learning model from scikit learn \cite{scikit-learn} that transforms a linear combination of features into probabilities using a logistic function, enabling classification into two classes based on these probabilities and a decision threshold. In order to simplify the problem to two classes, with class 1 for label 'a' and class -1 for labels 'b', 'c', and 'd'. We chose this model because it seemed the most suitable. We tried several approaches for the training set. The first method consists only in the test data from Subtask A, the second one involves both Test and Train data from Subtask A and the third approach comprises all the data from Substask A and from Experts.

%minimum document frequency set to one, no limit on maximum features, unicode strip accents, minimum number of documents set to one, no  enabled the use of inverse document frequency (IDF) reweighting, applied smoothing to the IDF weights, and implemented sublinear scaling to term frequency.
%\newline Other pre-processing techniques were performed using regular expressions in order to clean the text by removing special characters.
\begin{table*}
\centering
%\resizebox{\linewidth}{!}{%
\begin{tabular}{lcccc}
\textbf{submission} &
  \multicolumn{1}{l}{\textbf{recall}} &
  \multicolumn{1}{l}{\textbf{precision}} &
  \multicolumn{1}{l}{\textbf{recall\_w}} &
  \multicolumn{1}{l}{\textbf{harmonic}} \\ 
Test$^{1}$                              & 0.921 & 0.888 & 0.513 & 0.904 \\
Test + LLM$^{2}$                    & \textbf{0.939} & 0.890 & 0.390 & 0.914 \\ 
LLM$^{3}$                & 0.935 & \textbf{0.905} & \textbf{0.553} & \textbf{0.919} \\
\hline
TaskA\_3.1 + LLM   & 0.919 & 0.891 & 0.560 & 0.905 \\
A+E\_3.1 + LLM & 0.918 & 0.892 & \textbf{ 0.578} & 0.905 \\
LLM duplicates             & \textbf{0.941} & \textbf{0.907} & 0.398 & \textbf{0.924} \\
\hline
UoS NLP  & 0.943 &	\textbf{0.916} &	0.527 &	\textbf{0.929} \\
sophiaADS & \textbf{0.944} &	0.906 &	0.489 &	0.924 \\
UZH\_CLyp & 0.910 &	\textbf{0.916} &	\textbf{0.742} &	0.913 
\end{tabular}%
%}
\caption{Highlights evaluation scores of our systems in comparison to other participants in the Shared Task. The first three rows marked with superscript are the official versions we submitted during competition. The next 3 are additional experiments with highlights 3.1 or without removing duplicates and overlaps from LLM output.
The last three rows are submissions from other participants.
%Test represents traditional machine learning techniques and extractive summarization as described in Section \ref{sec:goml}; GOML+LLM highlights sentences of important features and produces summaries with the LLM; LLM duplicates is similar to the plain LLM prompting (Section \ref{sec:llm}) without ignoring duplicate highlights; TaskA and A+E are the same logistic regression as in GOML but trained on more data.
}
\label{tab:high}
\end{table*}

\begin{table}
\centering
\resizebox{\linewidth}{!}{%
\begin{tabular}{lcc}
\textbf{submission} & \multicolumn{1}{l}{\textbf{consistency}} & \multicolumn{1}{l}{\textbf{contradiction}} \\ 
Test$^{1}$                             & 0.901 &  0.238 \\
Test + LLM$^{2}$                    & \textbf{0.973} & \textbf{0.081} \\
LLM$^{3}$ & 0.964 & 0.104 \\ 
\hline
TaskA\_3.1   & 0.910 & 0.217 \\
A+E\_3.1 & 0.908 & 0.218 \\ %\cline{1-1}
TaskA\_3.1 + LLM    & 0.971 & 0.085 \\
A+E\_3.1 + LLM & \textbf{0.974} & \textbf{0.076} \\ %\cline{1-1}

\hline
UoS NLP & 0.966 &	0.107 \\
sophiaADS &  0.944 &	0.175 \\
UZH\_CLyp & \textbf{0.979} &	\textbf{0.064}
\end{tabular}%
}
\caption{Summary evaluation scores of our systems in comparison to other participants in the Shared Task. %The first three rows marked with superscript are the official versions we submitted during competition. %The next 3 are additional experiments with highlights 3.1 or without removing duplicates and overlaps from LLM output. The last three rows are submissions from other participants. %GOML represents traditional machine learning techniques and extractive summarization as described in Section \ref{sec:goml}; GOML+LLM highlights sentences of important features and produces summaries with the LLM; LLM duplicates is similar to the plain LLM prompting (Section \ref{sec:llm}) without ignoring duplicate highlights; TaskA and All are the same logistic regression as in GOML but trained on more data.
}
\label{tab:summary}
\end{table}

\textbf{2. SHAP} SHapley Additive exPlanations \cite{NIPS2017_7062} is an explainability library that implements several techniques to attribute individual contributions of each feature to a classifier's prediction. In our case, we use a simple linear explainer that assumes feature independence and ranks features based on a score computed as: $s_i = w_i (x_i - \hat{m}_i)$, where $w_i$ is the classifier coefficient of feature $i$, $x_i$ is the feature value in a post and $\hat{m}_i$ the mean of the feature value across all posts.

%various  We applied SHAP to the model's output to interpret it and extract the most important features that contribute to the assigned label.

\textbf{3. Selecting the highlights} requires matching the tokenized features from our tf-idf extractor to the text. We do so by aligning the different tokenizations using the Natural Language Toolkit \cite{bird2009natural} and retrieving the original verbatim strings. For highlight selection, we test \textbf{option 3.1} - highlights consisting of a context window of 14 words before and after each matched feature, not exceeding the sentence boundary. And \textbf{option 3.2} highlights consisting of entire sentences where important features are discovered in the original text.

\textbf{4. The summarization} consists of two options: \textbf{4.1} take the sentences found previously in step 3.2 and use an extractive summarization technique such as TextRank \cite{mihalcea-tarau-2004-textrank,PyTextRank} to generate a summary. This method is the fastest, but performed relatively poorly, obtaining high contradiction rates (0.238) and relatively low mean consistency (0.901). \textbf{Option 4.2 GOML+LLM} achieved the best overall performance and requires taking the sentences found previously and prompting a language model to generate an abstractive summary. Our best performing system in the official ranking is configured with option 3.2 (to extract full sentences as highlights) and option 4.2 (to generate summaries using LLM).

\section{Language Models}
\label{sec:llm}
For efficient text generation, we use a 4-bit quantized model (Q4\_K\_M) together with llama-cpp\footnote{\url{https://github.com/ggerganov/llama.cpp}} and langchain \cite{langchain2022} libraries. 
We use OpenHermes 2.5 based on Mistral \cite{Jiang2023Mistral7} that has been fine-tuned on code. According to the authors\footnote{\url{https://huggingface.co/teknium/OpenHermes-2.5-Mistral-7B}} training on a good ratio of code instruction of around 7-14\% of the total dataset boosted several noncode benchmarks, including TruthfulQA, AGIEval, and GPT4All suite.  
The language models approach can be summarized in the following steps:
\begin{enumerate}[label=(\alph*),topsep=0pt]
    \setlength\itemsep{-.11em}
    \item prompt the model using langchain to extract highlights from the texts for a number of $K=8$ times
    \item parse the LLM output and extract highlights from between quotation marks
    \item post-process responses: ensure the highlights are actually in the texts, remove duplicates, keep the longest matching highlights
    \item concatenate all posts and prompt the model without langchain to do a summary analysis of maximum 300 words
\end{enumerate}

Text generation parameters are set to a temperature of 0.75, top-p nucleus sampling 1, and a maximum context size of 32000.
To obtain as much data as possible, the LLM was run eight times on each post. The langchain prompt for extracting highlights is: \textit{Provide sequences of text that indicate that this person is suicidal? \textbackslash n \textbackslash n Post Body: \{post\_body\}}.
Each response is saved and post-processed to extract valid highlights present in the text, to remove duplicates, and to preserve the longest matching highlight. The model tends to be more verbose, no matter how much we change the prompt, so the post-processing step proved to be essential.

To extract summaries, we run the model only once with the following prompt: \textit{As a psychologist and expert therapist, summarize the content by identifying any indications of suicidal thoughts. Provide evidence from the text to support your analysis. \textbackslash n \textbackslash n Post Body: \{content\_body\}\textbackslash n \textbackslash n Analysis:}. When using GOML with Option 4.2, the content body consists in the concatenation of important sentences instead of the post bodies. We found that the model tends to hallucinate and copy paste content from the text, unless the word \textit{Analysis} is explicitly mentioned at the end.

%OpenHermes was trained on 1,000,000 entries of data generated primarily by GPT-4, as well as other high-quality data from open datasets across the AI landscape. Filtering was extensive of these public datasets, as well as conversion of all formats to ShareGPT, which was then further transformed by axolotl to use ChatML. These particular parameters influence the behavior of the language model during text generation: T

%TheBloke/OpenHermes-2.5-Mistral-7B-GGUF\footnote{https://huggingface.co/TheBloke/OpenHermes-2.5-Mistral-7B-GGUF}

%\newline We use langchain \cite{langchain2022} for generating the highlights with the prompt constructed specifically for a psychologist or expert therapist. The question requests a summary of the content with a focus on identifying any indications of suicidal thoughts from the content of the post and for generating the summary evidence of the content of the posts because the question template is designed for a more general analysis. 

\section{Results and Discussion}
Our three official submissions for the Shared Task in this order are:

\begin{itemize}[topsep=0pt]
    \setlength\itemsep{-.11em}
    \item Test$^{1}$ - GOML fit on the Task A test set (1.1), highlights consisting of a 14 word context window (3.1), and extractive summaries generated from important sentences (4.1)
    \item Test + LLM$^{2}$ - [our best submission] GOML fit on the Task A test set (1.1),  highlights consisting of entire sentences with important features (3.2), and LLM-generated abstractive summaries from combined sentences (4.2)
    \item LLM$^{3}$ - pipeline as described in \autoref{sec:llm}
\end{itemize}

%The organizers use two distinct metrics to evaluate the selection of highlights and the generation of summaries. 
Recall is computed as the average of the maximal semantic similarity between each gold highlight and all predicted highlights based on BERTScore \cite{zhang2019bertscore}. A point of critique that we can raise here is that introducing duplicate highlights of different sizes will generate a better overall recall score. In practice, such a system could potentially slow down an expert looking for indicators of suicide. For example, our submission "LLM duplicates" from \autoref{tab:high} does not remove highlights extracted from multiple runs of the LLM that are substrings of each other, and therefore obtains the highest recall. Similarly, systems that have shorter highlights (such as those that use the context around important features) achieve a lower recall than systems that return entire sentences as highlights. We do not know whether this is an artifact of BERTScore or from the way the annotations have been created. For example, the sophiaADS team \cite{yusuke2024} returns complete sentences using a fine-tuned BERT model and their method obtains the highest recall score in the competition. In both their case and ours, we can observe that the weighted recall penalizes results in which highlights are entire sentences.

For this downstream task of identifying highlights, we did not observe significant improvements in performance when training the logistic regression classifier with more data, nor did we observe a degradation of performance when training on the smallest amount of samples consisting only of the test set of Task A. This is encouraging for potential extensions of the GOML methodology to less-resourced languages.

The generated summaries are evaluated by taking the probability scores (from an external NLI tool) of having a summary that contradicts the gold sentence as a premise. In terms of consistency and contradiction \autoref{tab:summary}, the best results were obtained by Test + LLM$^{2}$ which combines the efficacy of extracting highlights of high recall (albeit low precision) with the ability of LLMs to generate adequate and coherent summary content. This is confirmed by the additional results combining LLM with GOML + option 3.1 with shorter summaries (\autoref{tab:high} rows four and five). These models achieve the highest consistency (.974) and lowest contradiction scores (.076) of our systems. Team UZH\_CLyp \cite{uluslu2024} uses retrieval augmented generation and provides additional context to the model when generating the summary to obtain the best results in the competition (given this criterion). This corroborates our observations that giving more concise or more focused content to LLMs leads to better generated summaries than providing the complete (and possibly noisy) post bodies from users to the LLM. 
The results of the team UoS NLP \cite{singhetal2024} are relatively similar to our LLM submissions that use chain-of-thought prompting to extract highlights and remove duplicates. Their LLM is based on Mixtral model quantized to 8 bits, which might explain the slight increase in evaluation scores across different metrics. 

While GOML performs competitively to more resource-intensive approaches in detecting highlights, the same cannot be said about summaries. Our Test$^{1}$ model that used TextRank for extractive summarization obtained one of the worst contradiction and consistency scores in the entire competition. Its main advantage remains that it can run the entire machine learning pipeline to train the classifier and generate all the evidence (highlights and summaries) for the 125 users in less than 60 seconds. In contrast, our quantized LLM on CPU runs in 3.5 hours for the same set of users. To be consistent with our comparisons, in all of our approaches, we have only used a CPU server with 7 cores and 64 GB of memory to compute the results.

Given the surprising efficacy of the traditional machine learning model, we ask whether sentences containing important features have specific linguistic characteristics. Sentences are divided into two categories: \textbf{important} if they contain important features for classification and with the label \textbf{other} otherwise. Our statistical analyses visible also in \autoref{fig:violins} indicate that important sentences are generally more likely to have pronouns, verbs, and adjectives. In terms of mean value, pronouns and verbs are statistically different at a p-value $< 0.05$ in important sentences more often than in the rest. Similarly, mean sentence lengths are statistically larger in important sentences than in the other ones.
Adverbs show no difference between the two classes, and adjectives and nouns obtain a p-value of $0.6$ after 100,000 permutations.  Given the nature of permutation tests, this is equivalent to saying that there is a 6\% chance of observing a difference in means for adjectives and nouns due to chance. 

Our brief analyses show that important sentences have different (statistically significant) linguistic patterns that can distinguish them from the rest. We believe that this could be one of the reasons behind the good evaluation scores and the suitability of the GOML approach to extract highlights from this particular dataset.

\section{Conclusions}
To conclude, our results show that a classifier paired with a machine learning explainability method can be a useful tool for identifying important sentences, phrases, and highlights that are representative of a given class. This is encouraging for languages where current LLMs do not perform as well or where the amount of data and compute resources is limited. Additionally, our experiments show that noisy generated output containing duplicates achieves better recall, leading to the conclusion that relying on a single metric can be detrimental to this task. We believe that ultimately expert human judgments would be the best measure for evaluating and selecting the most useful systems based on multiple criteria. 

In general, when investigating the output of LLM-based approaches, we could observe better quality in terms of the generated text and langchain reasoning. Our work shows that these results can be further improved by combining LLMs with good old-fashioned machine learning methods.

\section{Ethics}
Working with user posts that talk about inflicting self-harm is a difficult endeavor. Although our methods bring about a small contribution in the interdisciplinary field of suicidology, we must recognize that technological solutions are not always helpful in an impactful way for people who suffer. Our work was carried out with the greatest care for the privacy and management of this data. During human analyses, repeated exposure to suicide-related content can be triggering and potentially harmful. The authors have double-checked each other on their mental health and ability to work during the entire time of doing this work.

\section{Limitations}
\begin{itemize}
    \item Preserving duplicates or generating too many highlights can lead to an artificial increase in recall. The score increase can be misleading, since such a system can generate duplicates that are hard to interpret and not user-friendly. 
    \item LLM-generated summaries may include sexist biases, we have not observed these in a systematic manner, but on occasion the LLM would assign gendered pronouns to users who did not explicitly mention this in their posts. Further research is required to integrate multi-modal variables such as class, race, gender in the prediction mechanism.
    \item The data that we have to work with had strong signals of suicide risk, therefore, we wonder whether such an approach would still be suitable in cases where the linguistic signal is more subtle or whether our models are able to generalize on out-of-domain data.  
\end{itemize}

\section*{Acknowledgements}
We acknowledge the assistance of the American Association of Suicidology in making the dataset available. This work is supported by the Faculty of Mathematics and Computer Science, University of Bucharest, as part of the Archaeology of Intelligent Machines course.

%\begin{itemize}
%    \item We confirm that we have read \cite{benton-etal-2017-ethical} and that in our use of this dataset we are committed to maintaining that paper’s broad ethical principles.
%    \item We commit to citing the "University of Maryland Reddit Suicidality Dataset" \cite{shing-etal-2018-expert, zirikly-etal-2019-clpsych} in any publications using or discussing this dataset using including appropriate references (you can find BibTeX entries on the datasetvweb page):
%\end{itemize}

\bibliography{acl_latex}

\appendix
\section{Appendix}
\label{sec:appendix}

The first text classification training scenario involves only the test set from Task A, because it is the smallest (186 posts), and one should expect it to generate the weakest classifier. We gradually increase the data to see whether there are changes in the results by adding the entire Task A data (labeled in the results section as "TaskA"). Lastly, we include the entire Task A and expert data, referred to in the results section as "A+E". 

When running the tf-idf vectorizer we set the minimum document frequency to one, no limit on maximum features, Unicode strip accents, minimum number of documents set to one, enable the use of inverse document frequency (IDF) reweighting, smoothing to the IDF weights, and sublinear scaling to term frequency.

Logistic regression is set with balanced class weight, and we do not perform any hyperparameter optimization. Nevertheless, classifiers tend to predict only the majority class \autoref{tab:classification}, so the balanced accuracy score never increases significantly, regardless of the fold or amount of data used.

\begin{table}
\centering
\begin{tabular}{cccc}
%\hline
\textbf{Approach} & \textbf{Bal. Acc} & \textbf{Acc}  & \textbf{F1} \\
%\hline
test $\rightarrow$ Test     & .5      & .82   & .74    \\
+train $\rightarrow$ TaskA    & .5      & .82   & .74  \\
+expert $\rightarrow$ A+E   & .5      & .86   & .8    \\\hline
\end{tabular}
\caption{Stratified 5-fold cross-validation for binary risk prediction on different subsets of Task A and expert data. The first row represents cross-validation only on the test set, the second row adds the training set over the test set thus using the entire Task A, and the third row adds the expert data over all the previous. All values can vary between $\pm .05$ at different random shuffles.}
\label{tab:classification}
\end{table}

\begin{figure*}[htb]
    \centering
    \includegraphics[width=.8\linewidth]{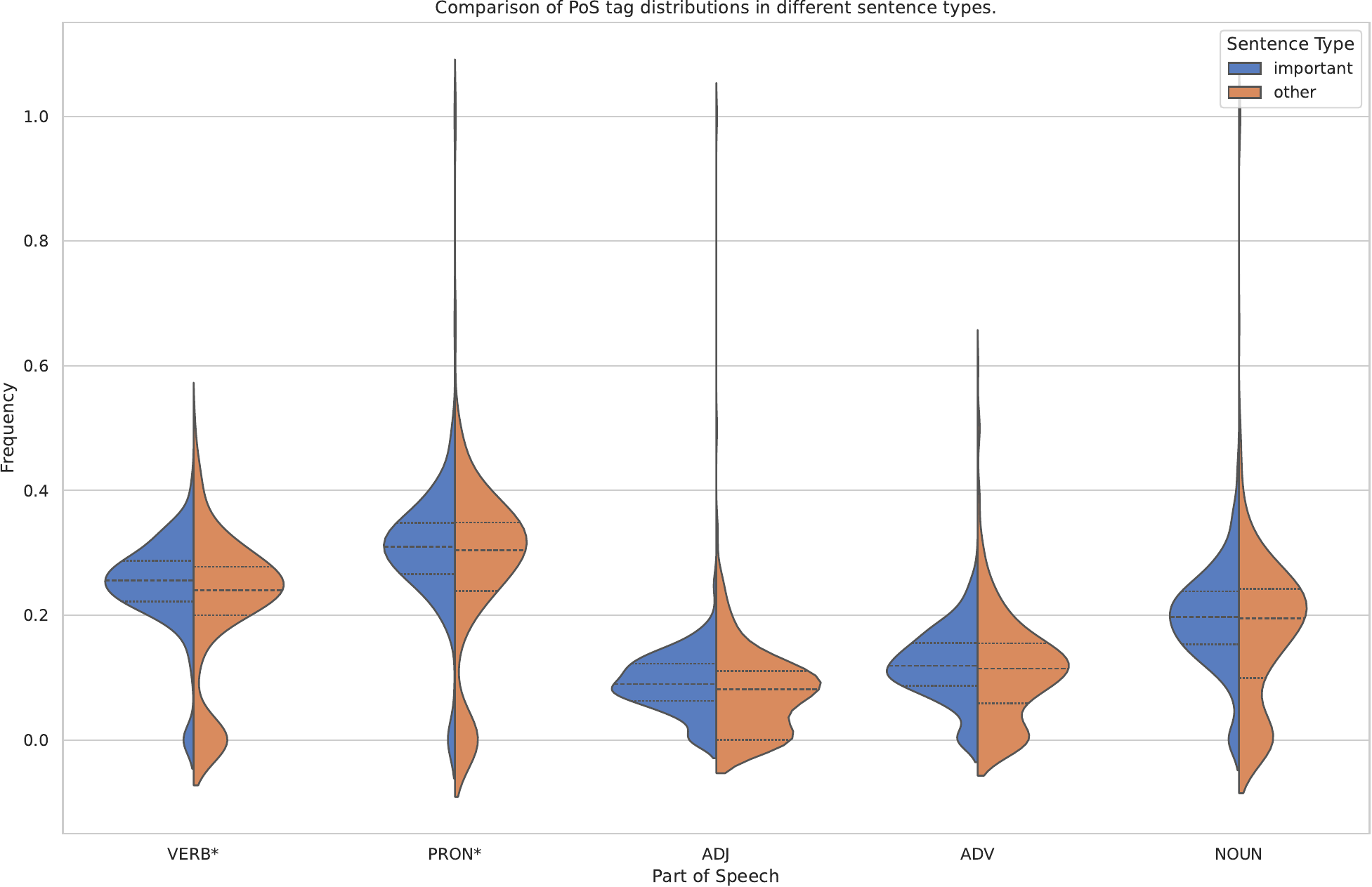}
    \caption{PoS tag distributions in sentences containing highlighted features (important) vs. other. Marked with * are PoS tags that have statistically significant means in a bootstrap permutation test at a p-value of $0.05$.}
    \label{fig:violins}
\end{figure*}

%that can easily be extended to less resourced languages by leveraging 

\section{What did Not Work}

\begin{itemize}
    \item Fine-tuning a LLM for classification with LoRA and unsloth library \footnote{\url{https://github.com/unslothai/unsloth}} using mistral-7b-bnb-4bit quantized model to classify the suicide risk by responding verbally; we were hoping to guide the model's attention towards important features for generating the content; after fine-tuning, the model was not able to produce good highlights.
    \item Given that OpenHermes 2.5 is fine-tuned on code, we were expecting to use grammars\footnote{\url{https://github.com/ggerganov/llama.cpp/blob/master/grammars}} to constrain the generation of highlights in the form of a list of strings, but the model proved not to perform very well in some of our empirical small-scale tests and we eventually abandoned this direction.
    \item We also tried to use Yake \cite{campos2020yake} to extract keywords from the titles and posts and then use this list of words as a parameter in TF-IDF. This approach did not work well because the list of extracted important features was too limited.
\end{itemize}

\end{document}